%% file: root.tex
\documentclass[letterpaper, 10 pt, conference]{IEEEtran}  

\IEEEoverridecommandlockouts                              

\usepackage{graphics} 
\usepackage{epsfig} 
\usepackage{mathptmx} 
\usepackage{times} 
\usepackage{amsmath} 
\usepackage{amssymb}  
\usepackage{soul}
\usepackage{lipsum}
\usepackage[binary-units=true,per-mode=symbol]{siunitx}
\usepackage{epstopdf,placeins}
\usepackage[compress]{cite}
\usepackage{siunitx}
\usepackage{tikz}
\usepackage{xltabular}
\usepackage{booktabs}
\usepackage{makecell}
\usepackage{float}
\usepackage{graphicx}
\usepackage{multirow}
\usepackage{pgfplots}
\usepackage{array}
\usepackage{tabularray}
\usepackage{boldline}
\usepackage[hyphens]{url}
\usepackage[bottom]{footmisc}
\usetikzlibrary{arrows,chains,positioning,fit}
\usepackage{marvosym}
\usepackage[a4paper, left=0.625in, right=0.625in, bottom=1in, top=0.75in]{geometry}

\usepackage[nolist]{acronym} 
\usepackage{hyperref} 

\usepackage[colorinlistoftodos, textsize=small, textwidth=50mm]{todonotes}
\title{Survey on Teleoperation Concepts for Automated Vehicles}

\author{Domagoj Majstorovi\'c, Simon Hoffmann, Florian Pfab, Andreas Schimpe,\\ Maria-Magdalena Wolf and Frank Diermeyer
\thanks{The authors are with the Institute of Automotive Technology at the Technical University of Munich (TUM), 85748 Garching bei M\"unchen, Germany. Email: {\tt\small \{firstname\}.\{lastname\}@tum.de}}%
}

\begin{document}
\maketitle
\thispagestyle{empty}
\pagestyle{empty}

\input{Chapters/Acronym.tex}
\input{Chapters/Abstract.tex}
\input{Chapters/Introduction.tex}
\input{Chapters/Survey.tex}
\input{Chapters/Patents.tex}
\input{Chapters/Conclusion.tex}
\input{Chapters/Acknowledgements.tex}

\bibliographystyle{IEEEtranBST/IEEEtran}
\bibliography{IEEEtranBST/IEEEabrv,Literature}

\end{document}

%% file: Chapters/Acronym.tex
\begin{acronym}
%
%

\acro{av}[AV]{Automated Vehicle}
\acroplural{av}[AVs]{Automated Vehicles}


\acro{ad}[AD]{Automated Driving}

\acro{ads}[ADS]{Automated Driving System}
\acroplural{ads}[ADS]{Automated Driving Systems}
\acro{ugv}[UGV]{Unmanned Ground Vehicle}
\acroplural{ugv}[UGVs]{Unmanned Ground Vehicles}
\acro{rrt}[RRT]{Rapidly-Exploring Random Tree}
\acro{cots}[COTS]{Commercial Off-The-Shelf}

\acro{odd}[ODD]{Operational Design Domain}
%
%
%
%
%
%

\acro{hmd}[HMD]{Head-Mounted Display}
\acroplural{hmd}[HMDs]{Head-Mounted Displays}

%
%

\acro{hmi}[HMI]{Human-Machine Interface}


\acro{orccad}[ORCCAD]{Open Robot Controller Computer-Aided Design}

\acro{swa}[SWA]{Steering Wheel Angle}

%

\acro{ecs}[ECS]{Entity Component System}
\acro{ros}[ROS]{Robot Operating System}
\acro{rwa}[RWA]{Road Wheel Angle}


\acro{mpc}[MPC]{Model Predictive Control}


\acro{tod}[ToD]{Teleoperated Driving}
\acro{gui}[GUI]{Graphical User Interface}
\acro{g2g}[G2G]{Glass-to-Glass}

\end{acronym}

%% file: Chapters/Abstract.tex
\begin{abstract}
In parallel with the advancement of Automated Driving~(AD) functions, teleoperation has grown in popularity over recent years. 
By enabling remote operation of automated vehicles, teleoperation can be established as a reliable fallback solution for operational design domain limits and edge cases of AD functions. 
Over the years, a variety of different teleoperation concepts as to how a human operator can remotely support or substitute an AD function have been proposed in the literature. 
This paper presents the results of a literature survey on teleoperation concepts for road vehicles. Furthermore, due to the increasing interest within the industry, insights on patents and overall company activities in the field of teleoperation are presented.
\end{abstract}

%% file: Chapters/Introduction.tex
\section{\uppercase{Introduction}}
\label{sec:introduction}
\acp{av} are the key technology of tomorrow's mobility.
Their development has been a focus of academic and industrial research for decades, and market introduction is increasingly imminent.
The first truly driverless vehicles, i.e., without a safety driver, are already to be found on public roads in different parts of the world.
However, remaining challenges and complex or even currently unknown edge cases pose a real threat to the utilization possibilities of \acp{av}.
With increasing popularity, a teleoperation technology is seen as and being developed as a fallback solution for \ac{ad} functions.
By enabling remote operation, the \ac{av} can be supported by a human operator who is not located in the vehicle. 
Requesting this remote support whenever an \ac{ad} function reaches the limits of its \ac{odd}, the objective of teleoperation technology is to provide a safe and efficient solution to overcome these limitations. 
Once the \ac{av} is brought back into its nominal \ac{odd}, it can again continue its journey fully automated as before.
Over the years, a variety of different teleoperation concepts relating to how a human operator can remotely support or substitute an \ac{ad} function have been proposed in the literature. 
This paper presents the results of a literature survey on teleoperation concepts for road vehicles and provides insights into relevant company activities and patents in the field of \ac{av} teleoperation.
\subsection{Human Interaction with Automated Vehicles}
\label{sec:relatedwork}
The idea that a human could assist the \ac{av} on the fly in situations where the vehicle is uncertain about what to do, can also be found outside the context of remote vehicle control.
Guo et al.~\cite{Guo2019hierarchicalCoopCtrl} proposed a hierarchical driver-vehicle cooperation framework that includes interfaces at different functional levels to incorporate actions from an on-board human passenger into the actions of the \ac{av}.


Further to this, Walch \cite{Walch2021}, in his driving simulator studies has demonstrated in different use cases that such human-machine collaboration with different input modalities could be a feasible way of overcoming \ac{av} system weaknesses while still enjoying a high level of acceptance among users.

However, while these concepts offer valuable insights into modes of human interaction with automated vehicles, they do not utilize teleoperation technology and as such were not considered in the survey presented in this paper.



%
%
\subsection{Taxonomy}
Throughout the years, various terms have been used to describe similar or even the same teleoperation techniques. 
With increasing activity in this field, this became even more apparent. 
Bogdoll et al.~\cite{Bogdoll2021} identified this lack of common taxonomy for teleoperation concepts and gave an extensive overview of the terminology used across the automotive domain, while proposing improvements. 
Building upon that taxonomy survey, the SAE J3016:2021 standard~\cite{SAE2021} and guidelines on automated vehicles \cite{CentreforConnected&AutomatedVehicles2022, CentreforConnectedandAutonomousVehicles2020}, Table~\ref{tab:taxonomy} shows the adopted terminology used in this paper. The term teleoperation is used as an umbrella term that encompasses all functionalities needed to remotely support operation of automated vehicles. 

\input{Tables/taxonomy_table_v2.tex}
\input{Figures/conceptsOverview}
\subsection{Contributions}
Over the years, different teleoperation concepts for road vehicles have been developed and researched.
However, until now, have not been systematically summarized and categorized.
In this work, these concepts are collected and grouped in respect of their technical functionalities. 
Their advantages and disadvantages are highlighted, emphasizing the research motivation and discussing respective industry activities. 
Finally, this work contributes to establish existing taxonomy in the field while making proposals for the concepts not introduced or covered so far. \\

%% file: Tables/taxonomy_table_v2.tex
\begin{table}[b]
	\caption{Adopted Taxonomy}
	\label{tab:taxonomy}
	\centering
	\begin{tblr}{
			stretch=1.25,
			colspec={|Q[c,0.5cm]Q[r,1.75cm]|Q[l,4.5cm]|},
			rowspec={Q[m]Q[m]Q[m]Q[m]},
			vline{2,2} = {2-4}{1pt,solid},
			vline{1,1} = {1-7}{1pt},
			vline{4,1} = {1-7}{1pt}}
		\hline[1pt]
		& \textbf{Term}  & \textbf{Description}\\
		\hline[1pt]
		\SetCell[r=3,c=1]{c} \rotatebox[origin=c]{90}{\textsc{Teleoperation}}
		& \textsc{Remote Driving} & System is fully under remote control. \\ \hline
		& \textsc{Remote Assistance}  & System receives event-driven remote assistance from the operator, while still being responsible for the driving task. \\ \hline
		& \textsc{Remote Monitoring}  & System is remotely monitored with very limited intervention possibilities. \\ \hline[1pt]
	\end{tblr}
\end{table}

%% file: Figures/conceptsOverview.tex
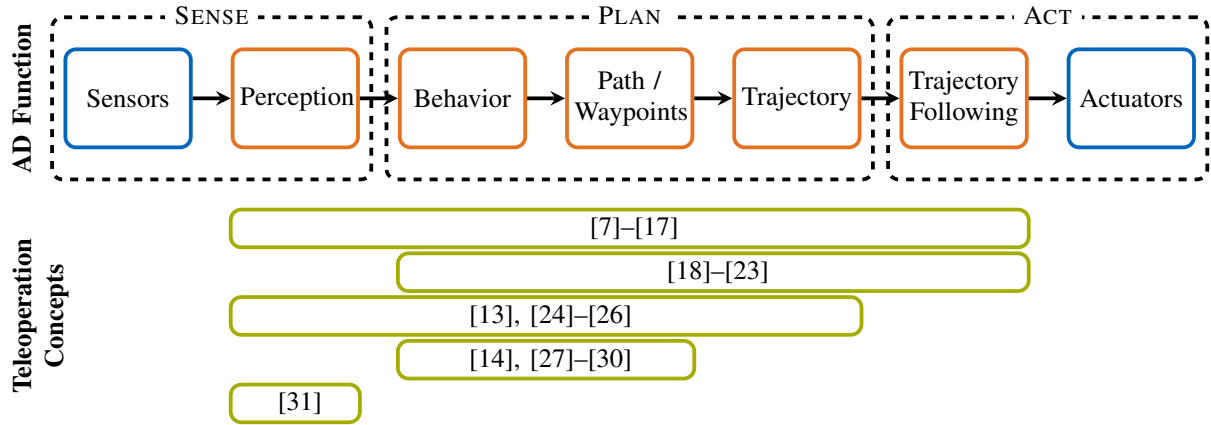
\begin{figure*}[t]
\centering
\begin{tikzpicture}[node distance = 5mm, on grid]

\definecolor{tumBlau}{RGB}{0,101,189}
\definecolor{schwarz}{RGB}{0,0,0}
\definecolor{tumOrange}{RGB}{227,114,34}
\definecolor{tumGruen}{RGB}{164,173,0}

\def\lineWidth{0.5mm}

\def\barHeight{0.5cm}
\def\offsetFirstBar{0.8cm}
\def\offsetBars{0.08cm}

\def\textwidthNodes{1.45cm}
\def\innerSepNodes{0.1cm}

\def\innerXSep{0.15cm}
\def\innerYSep{0.4cm}

\newcommand{\newBar}[5]{
	\def\offsetCurr{-\offsetFirstBar - #1*(\barHeight+\offsetBars)}
	\node[draw = tumGruen, line width=\lineWidth, fit={([yshift=\offsetCurr]#2.south west) ([yshift=\offsetCurr -\barHeight]#3.south east)}, inner sep=0pt, rounded corners] (#5) {};
	\node at (#5.center) [] {\text{#4}};
}

\tikzset{
	hardware/.style={draw=tumBlau,rectangle,rounded corners, minimum height=1.3cm, align=center, text width=\textwidthNodes, inner sep=\innerSepNodes, line width=\lineWidth},
	software/.style={draw=tumOrange,rectangle,rounded corners, minimum height=1.3cm, align=center,text width=\textwidthNodes, inner sep=\innerSepNodes, line width=\lineWidth},
	%
	containerSPA/.style={draw, rectangle, rounded corners, dashed,inner ysep=\innerYSep,inner xsep=\innerXSep, minimum height=1cm, line width=\lineWidth},
	containerMain/.style={rectangle,rounded corners, inner ysep=\innerYSep, inner xsep=\innerXSep, minimum height=1cm, line width=\lineWidth},
	%
	arrow/.style={draw, line width=\lineWidth, ->, >=stealth},
	line/.style={draw, line width=\lineWidth, -, >=stealth},
}


\node [hardware] (sensors) {Sensors};
\node [software, right = of sensors.east] (perception) {Perception};
\node [software, right = of perception.east] (behaviour) {Behavior};
\node [software, right = of behaviour.east] (pathPlan) {Path / \\Waypoints};
\node [software, right = of pathPlan.east] (trajPlan) {Trajectory};
\node [software, right = of trajPlan.east] (trajFollow) {Trajectory Following};
\node [hardware, right = of trajFollow.east] (act) {Actuators};

\draw [arrow] (sensors.east) -- (perception.west); 
\draw [arrow] (perception.east) -- (behaviour.west); 
\draw [arrow] (behaviour.east) -- (pathPlan.west); 
\draw [arrow] (pathPlan.east) -- (trajPlan.west); 
\draw [arrow] (trajPlan.east) -- (trajFollow.west); 
\draw [arrow] (trajFollow.east) -- (act.west); 


\newBar{0}{perception}{trajFollow}{
	\cite{Bensoussan1997, Appelqvist2007, Gnatzig2013ToDSystemDesign, Shen2016, Hofbauer2020telecarla, Schimpe21oss4tod, Jatzkowski2021, 5GCroCoD2p2, Ross2007, Bodell2016, Wu}} {bar5};

\newBar{1}{behaviour}{trajFollow}{
	\cite{Anderson2013, Schimpe2020steerwithme, Qiao2021, Schitz2021acc, Storms2017sharedControl4oa, Saparia21ass4tod}} {bar4};

\newBar{2}{perception}{trajPlan}{
	\cite{Jatzkowski2021, Gnatzig2012trajBasedSharedAutonomy, Hoffmann2022a, Kay1995a}} {bar3};

\newBar{3}{behaviour}{pathPlan}{
	\cite{5GCroCoD2p2, Bjornberg2020, Schitz2021, Hosseini2014a, Schitz2021interactivePathPlanning}} {bar2};

\newBar{4}{perception}{perception}{\cite{Feiler2021percmod}}{bar1};

\begin{scope}[]
	\node [containerMain,fit=(bar2)(bar1)(bar3)(bar4)(bar5)] (Bars) {};
	\node at (Bars.west) [align=center, fill=white, rotate=90, yshift=2.3cm] {\textbf{Teleoperation}\\ \textbf{Concepts}};
\end{scope}

\begin{scope}[]
	\node [containerSPA,fit=(sensors)(perception)] (Sense) {};
	\node at (Sense.north) [fill=white] {\textsc{Sense}};
\end{scope}

\begin{scope}[]
	\node [containerSPA,fit=(behaviour)(pathPlan)(trajPlan)] (Plan) {};
	\node at (Plan.north) [fill=white] {\textsc{Plan}};
\end{scope}

\begin{scope}[]
	\node [containerSPA,fit=(act)(trajFollow)] (Act) {};
	\node at (Act.north) [fill=white] {\textsc{Act}};
\end{scope}

\begin{scope}[]
	\node [containerMain,fit=(Sense)(Plan)(Act)] (Automation) {};
	\node at (Automation.west) [fill=white, rotate=90, yshift=0.2cm] {\textbf{AD Function}};
\end{scope}

\end{tikzpicture}

\caption{An overview of reviewed teleoperation concepts. The upper part of the graphic shows the \ac{ad} functionalities, while the bottom part with its width and position depicts the level of remote control concept intervention with either a module interaction or complete replacement.} \label{fig:conceptOverview}
\end{figure*}

%% file: Chapters/Survey.tex
%
%
%
\section{\uppercase{Scientific Literature on Teleoperation Concepts}}
\label{sec:survey}
In this section, \ac{av} teleoperation concepts available in the literature are reviewed and grouped with respect to their functionalities.
This review includes remote driving and remote assistance teleoperation concepts, while remote monitoring has been omitted because of its limited possibilities for influencing \ac{av} operation\footnote{E.g. triggering emergency braking or a minimal risk maneuver with no possibility to influence the driving task or assist the vehicle while in nominal operation mode.}.
Fig.~\ref{fig:conceptOverview}, summarizes and graphically presents the reviewed teleoperation concepts. 
The upper part of the figure illustrates the simplified functionalities of an \ac{ad} function. 
This pipeline consists of the modules (1)~\textsc{Sense}, comprising sensors and perception algorithms, (2)~\textsc{Plan}ning of behavior, path and subsequent trajectory, and (3)~\textsc{Act} with trajectory following module and the vehicle actuators. 
The bottom of the figure depicts the reviewed remote control concepts as they relate to the \ac{ad} pipeline. 
The position and width of each bar correspond to the design of the teleoperation concepts proposed in the respective literature, meaning the concepts either replace the corresponding \ac{ad} function blocks or enable interaction with them.
Furthermore, in Table~\ref{tab:contrConcepts}, the approaches are grouped thematically according to similar technical functionalities. 
For each concept group, a denomination is defined. 
In the following subsections, these concept groups and their respective references are reviewed in more technical detail.

\input{Tables/overview_table_v2.tex}
\subsection{Direct Control}
\label{sec:dirCtrl}
\begin{figure}[b]
	\centering
	\includegraphics[width=80mm]{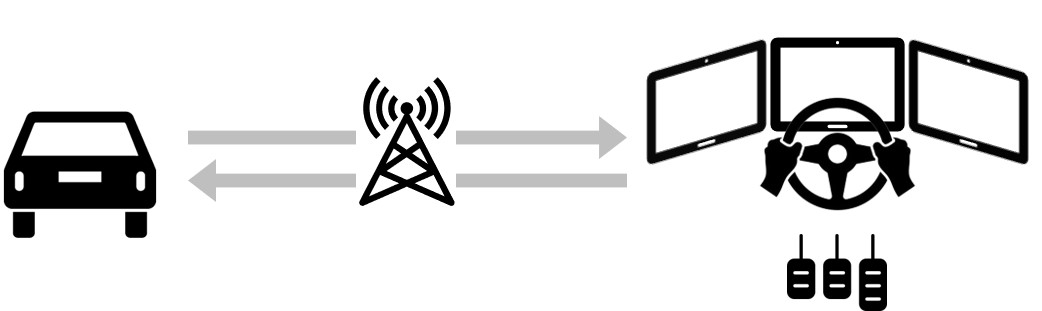}
	\caption[DirectControl]{Direct control concept.	A human operator observes received sensor data and directly provides control signals for the vehicle.}
	\label{fig:DirectControl}
\end{figure}

The most fundamental remote control concept is direct control, illustrated in Fig.~\ref{fig:DirectControl}. 
A mobile network is used to transmit sensor data from the \ac{av} to the operator. 
The data are visualized and the operator provides control signals, such as steering wheel angle, throttle and brake pedal position, gear shifts and various other events (e.g. horn, headlights, wipers), that are transmitted back to the \ac{av}. 
Direct control is subject to different challenges that the remote operator has to cope with, such as transmission latency~\cite{Georg2020} or reduced situational awareness~\cite{Mutzenich2021}.

In 1997, Bensoussan and Parent~\cite{Bensoussan1997} introduced a direct control concept that relies on visual information from a camera coupled with data from ultrasound sensors. 
The operator uses a joystick to interact with the vehicle. 
Even for this limited system, the advantage of the human operator, being in a safe and remote location while the vehicle operates in a potentially hostile or dangerous environment, was evident.
However, it also became apparent that transmission latency and situational awareness pose a real challenge in utilizing this control concept. 
Appelqvist et al.~\cite{Appelqvist2007} developed a remote control setup in which the teleoperation technology is mainly built from \ac{cots} components by realizing the direct control interaction between the operator and the vehicle through a steering wheel and pedals. 
In addition, the authors provided technical details on latency measurements as well as considerations on the overall system performance. 
In 2013, Gnatzig et al.~\cite{Gnatzig2013ToDSystemDesign} introduced and experimentally verified a general system design for teleoperated road vehicles.
This system relies on video streams coupled with occupancy grid map data obtained by laser sensors.
Additionally, Shen et al.~\cite{Shen2016} used a \ac{hmd} with the objective of improving the driving performance of the operator.

With the growing interest in remote operation systems, Hofbauer et al.~\cite{Hofbauer2020telecarla} developed a system enabling direct control interaction with a vehicle being teleoperated in the CARLA driving simulator. 
Schimpe et al.~\cite{Schimpe21oss4tod} tried to fill the gap in publicly available software for remote driving functionalities by publishing an open source software implementation. 
One objective of the software stack is to enable quick and flexible deployment to different types of automotive vehicles. 
Based on these research activities, this direct control implementation was successfully used within the projects UNICARagil~\cite{Jatzkowski2021} and 5GCroCo~\cite{5GCroCoD2p2}.

Over the years, various testbeds were developed to provide information on whether, when, and how direct control can be used. 
Research activities generally used passenger vehicles for experiments.
However, some other vehicle types also served as demonstrators. 
Ross et al.~\cite{Ross2007} investigated the minimum system requirements for effective teleoperation of an off-road ground combat vehicle. 
Bodell and Gulliksson~\cite{Bodell2016} remotely controlled a truck within a simulation environment and evaluated the handover process between autonomous operation and manual control.
Finally, Wu~\cite{Wu} applied the teleoperation to recreational vehicles by remotely controlling an electric golf cart. 

Due to technological progress in hardware and software, the direct control concept has reached a high degree of maturity and is the subject of numerous publications. 
However, challenges concerning the direct teleoperation concept still remain. 
The performance of the human operator as well as the safety of the vehicle is highly dependent on the latency~\cite{Georg2020} and stability of the mobile network~\cite{Hoffmann2021safetyAssessment}. 
Consequently, different concepts for further improvement of the direct control experience were investigated and implemented. 
Taking the transmission latency into account, Chucholowski et al.~\cite{Chucholowski2013a} presented a predictive display for the operator, where the position of the teleoperated vehicle and interacting dynamic objects are predicted.
Moreover, Tang et al.~\cite{Tang2014c} introduced the Free Corridor to reach a safe end state in the event of a connection failure. 
Graf et al.~\cite{Graf2020a} combined the benefits of these two concepts into the Predictive Corridor approach. 

Another critical challenge of direct control is the operator's reduced situational awareness by only perceiving the vehicle environment via various visualizations of sensor data~\cite{Mutzenich2021}. 
To improve the situational awareness, Georg et al.~\cite{Georg2019a} developed an adaptable interface for teleoperation and evaluated different visualization concepts~\cite{Georgb}, including the use of an \ac{hmd}~\cite{Georg2018}. 
\acp{hmd} were also investigated by Shen et al.~\cite{Shen2016} and Bout et al.~\cite{Bout2017}. 
Hosseini et al. took a different approach to compensate for the challenges mentioned above by introducing assistance systems for use in the execution of longitudinal~\cite{Hosseini2016e} and lateral~\cite{Hosseini2016a} guidance.
\subsection{Shared Control}
\label{sec:sharedCtrl}
\begin{figure}[b]
	\centering
	\includegraphics[width=80mm]{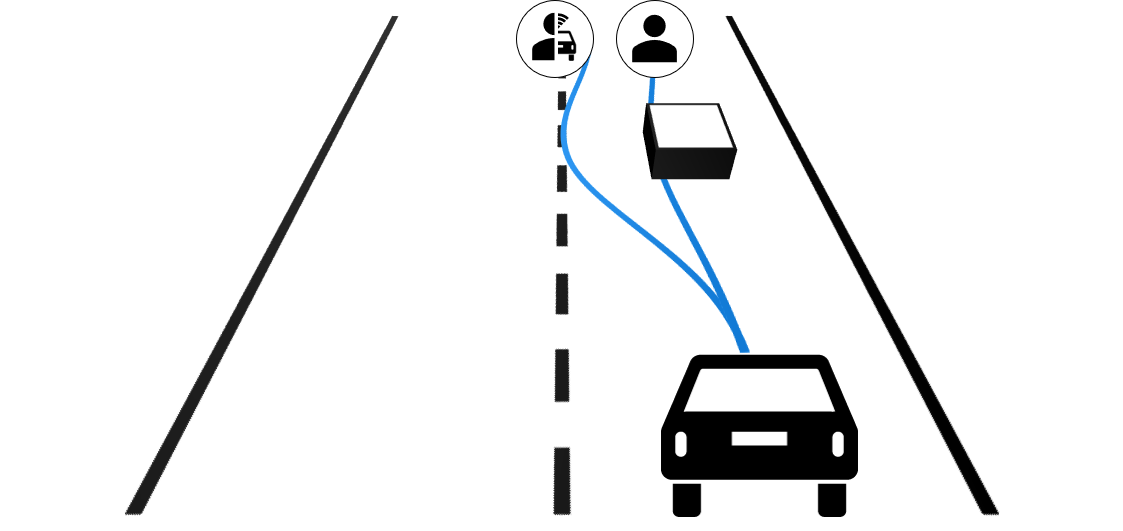}
	\caption[SharedControl]{Shared control concept. The controller can override the human operator in real-time in order to keep the ego vehicle and other road participants safe.}
\label{fig:SharedControl}
\end{figure}
In recent years, shared control emerged as a teleoperation concept that copes with the insufficiencies of the direct control concept. 
Its primary objective is to assist the operator in real-time to improve the safety of both the ego vehicle and other road participants. 
Of the \ac{ad} functions, only a functional perception module is required, i.e., the proposed approaches work with an object list or a representation of the free space as an input. 
As for direct control, the interface for the human operator is a steering wheel and throttle/brake pedals for lateral and longitudinal motion control commands respectively. 
The operator's control commands are transmitted to the vehicle, where they are accepted by the shared controller as desired control actions. 
If the shared controller deems them safe, the commands are executed. 
If not, the shared controller is able to intervene, i.e., override the control actions in order to prevent an imminent collision with an obstacle or road departure, as visualized in Fig.~\ref{fig:SharedControl}.

In the literature on  shared control, the control method \ac{mpc} is well-established, forming the basis for all approaches reviewed in this survey.
Anderson et al.~\cite{Anderson2013} developed an automation component, computing safe steering commands in a free corridor.
Based on a risk metric, the risk of the operator's steering commands is continuously evaluated, so that in situations where a greater risk is anticipated, the automation is granted a higher control authority. 
This approach was validated in a user study with a real test vehicle and human operators in the loop. 
Going beyond this, Schimpe and Diermeyer~\cite{Schimpe2020steerwithme} proposed an \ac{mpc}-based shared steering control approach for the avoidance of obstacles, which are modeled as repulsive potential fields. 
The corresponding validation was carried out in simulations without an actual human in the loop.
Another approach, suggested by Qiao et al.~\cite{Qiao2021}, models the human-machine interaction through Nash equilibrium-based, non-cooperative games.
Although the hardware of a real teleoperation test vehicle was introduced, the presented results are generated within a simulation environment.

Schitz et al.~\cite{Schitz2021acc} introduced an \ac{mpc}-based assistance approach in a cruise control (velocity-only) fashion. 
The controller maintains the vehicle at a safe velocity and distance to dynamic preceding vehicles and cross-traffic in urban scenarios. 
While experiments with a test vehicle are reported, details about the operator's interface and the teleoperation setup are not given.

Finally, two approaches are cited that are capable of overriding and combining both operator's steering actions as well as velocity control commands. 
First, Storms et al.~\cite{Storms2017sharedControl4oa} presented a shared control system based on \ac{mpc}, that assists the operator in the task of static obstacle avoidance. 
The controller was validated in studies by remotely controlling small mobile robots within an unstructured simulation environment with a human in the loop.
Second, Saparia et al.~\cite{Saparia21ass4tod} developed another \ac{mpc}-based shared control approach for vehicle teleoperation in urban scenarios.
Using a concept of \ac{mpc}-based predictive display, a strategy for mitigating latency in the teleoperation system was also proposed.
Validation of the controller was carried out in simulation.

In general, shared control is subject to similar challenges to those faced in direct control as described in the previous section. 
However, with the requirement of a functional perception module, the operator is assisted in terms of collision avoidance, effectively improving safety.  
\subsection{Trajectory Guidance}
\label{sec:trajectoryGuidance}
\begin{figure}[b]
	\centering
	\includegraphics[width=80mm]{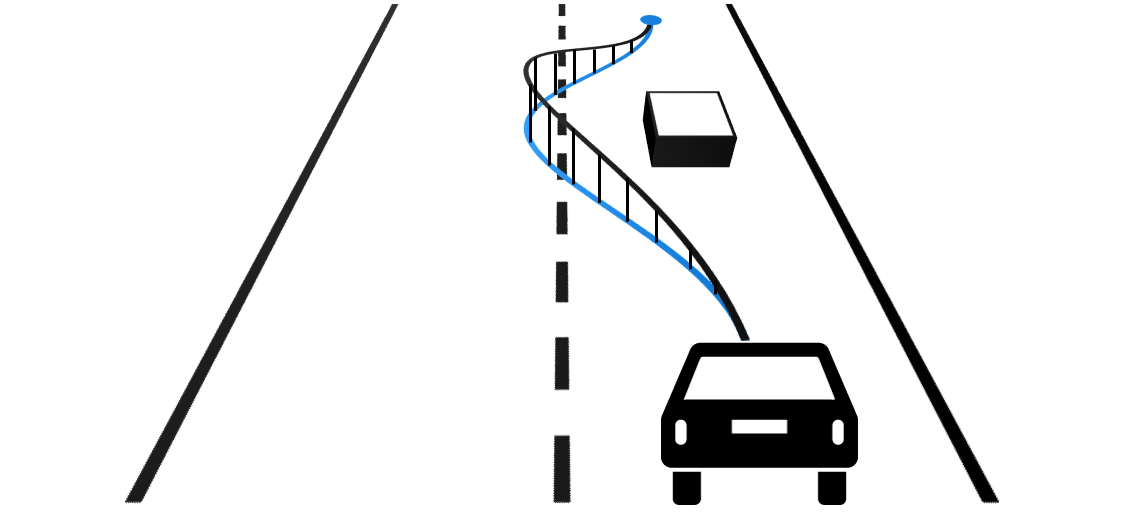}
	\caption[TrajGuid]{Trajectory Guidance concept. A human operator guides the vehicle through trajectories, taking over the tasks of perception and planning.}
	\label{fig:TrajGuid}
\end{figure}

In trajectory guidance mode the operator takes over the tasks of perception and the entirety of planning, which enables this concept to overcome problems at various levels of the \ac{ad} pipeline.
The operator's control commands are given as trajectories consisting of a path and corresponding velocity profile, as visualized in Fig.~\ref{fig:TrajGuid}. This trajectory is then tracked by the vehicle which is only responsible for the low-level control.
This relieves the operator to some extent, especially with regard to latency-critical control tasks.
Since the same interface from the trajectory planning module of the \ac{ad} function is used, remote trajectory guidance can easily be integrated within the \ac{av} software stack. 

Gnatzig et al.~\cite{Gnatzig2012trajBasedSharedAutonomy} implemented a trajectory guidance approach in a discrete manner.
The operator provides trajectory segments of several meters generated with a joystick or steering wheel and pedals. 
The vehicle follows this trajectory and stops at the end if the operator has not appended another trajectory segment in the meantime. 
While this teleoperation concept decouples the operator from the task of vehicle stabilization, it does not allow for dynamic adaptation of the current vehicle behavior.

Hoffmann et al.~\cite{Hoffmann2022a} proposed a concept that converts each control command provided through steering wheel and pedals, into a desired trajectory that ends in a standstill.
While this allows for dynamic adaptation of the desired vehicle behavior, the level of decoupling between the human and vehicle stabilization task is reduced because the human operator plans the desired trajectory at higher frequencies.


Jatzkowski et al.~\cite{Jatzkowski2021} also introduced a concept that provides trajectories for the trajectory-following controller of the \ac{ad} function pipeline.
However, further details on how these trajectories are specified were not given.
Kay and Thorpe~\cite{Kay1995a} developed a concept that allows the operator to define discrete waypoints that are associated with the desired velocity, effectively turning a waypoint sequence into a trajectory that is strictly followed by the vehicle.
This concept is closely related to the next subsection, with the main difference being the missing trajectory (re-)planning on the vehicle side.
%
%
\subsection{Waypoint Guidance}
\label{sec:waypointGuidance}
\begin{figure}[b]
	\centering
	\includegraphics[width=80mm]{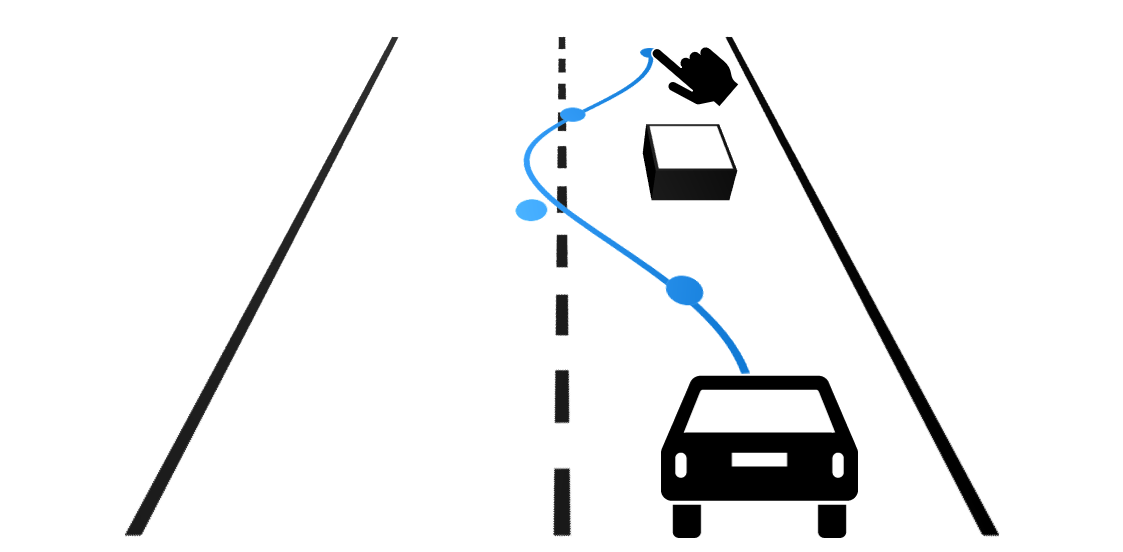}
	\caption[WaypointGuidance]{Waypoint guidance concept. A human operator specifies waypoints, taking over the tasks of decision making and path planning.}
	\label{fig:WaypointGuidance}
\end{figure}

Waypoint guidance is a teleoperation concept in which the human operator takes responsibility for decision making and path planning. 
As shown in Fig.~\ref{fig:WaypointGuidance}, the operator typically specifies discrete waypoints on a camera image or a map through mouse clicks. 
These waypoints are then fitted into a path that is transmitted to the vehicle, where the trajectory planning module of the \ac{ad} pipeline aims at tracking it, while taking into account  vehicle surroundings perceived by the perception module. 
As a result, the traveled path can deviate from the specified waypoints to some extent.

Approaches for waypoint guidance have been developed in different flavors. 
For instance, a basic implementation often referred to as indirect control, was demonstrated in the project 5GCroCo~\cite{5GCroCoD2p2}. 
Bj\"ornberg~\cite{Bjornberg2020}  created waypoints by driving a simulated vehicle in a virtual environment.
The real vehicle then mimics the behavior of the simulated vehicle by following the recorded waypoints.
Finally, the approach of Schitz et al.~\cite{Schitz2021} creates a corridor along mouse-clicked waypoints, in which the vehicle performs trajectory planning in an automated manner.

To conclude, with waypoint guidance the operator is not responsible for low-level control and the \ac{av} makes the final decision, so a larger part of the \ac{ad} pipeline must be functional as shown in Fig.~\ref{fig:conceptOverview}. This also means that only a limited dynamic intervention by the operator is possible, which might result in time-inefficient stop-and-go driving behavior. On the other hand, this property makes this control concept more robust in the face of network latency issues.
\subsection{Interactive Path Planning}
\label{sec:interactPathPlan}

\begin{figure}[!h]
	\centering
	\includegraphics[width=79mm]{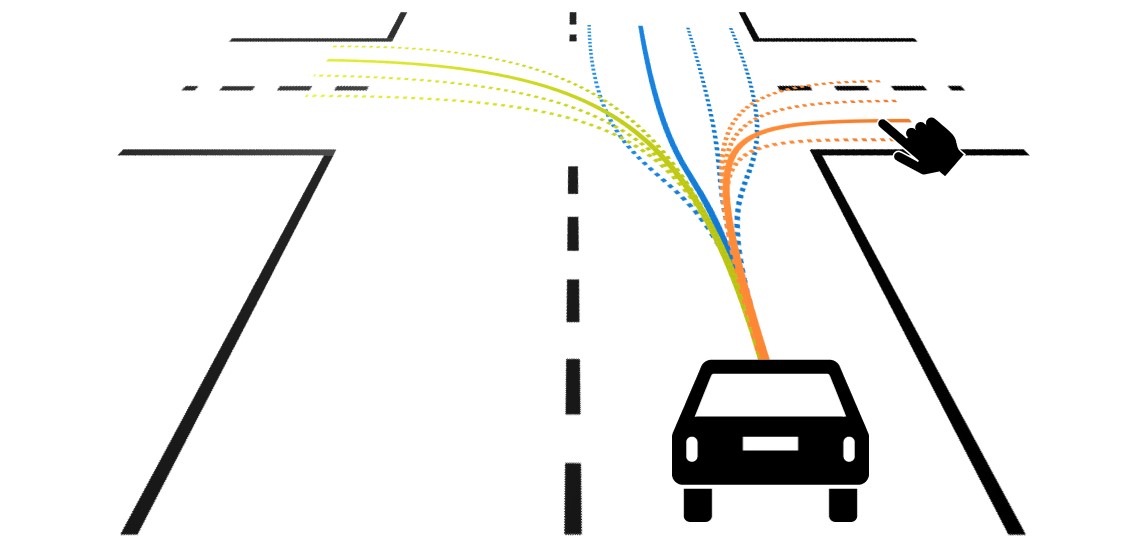}
	\caption[IPP]{Interactive path planning concept. A human operator takes over decision making and selects a path based on multiple suggestions.}
	\label{fig:IPP}
\end{figure}

Interactive path planning allows the operator to intervene at the decision making level of the \ac{ad} pipeline. 
This concept was introduced by Hosseini et al.~\cite{Hosseini2014a} by enabling real-time interaction between human and machine at the path planning level, and was later extended by Schitz et al.~\cite{Schitz2021interactivePathPlanning}. 
Optimal paths are computed automatically and visualized through augmented reality, where the operator chooses a suggested path, as shown in Fig.~\ref{fig:IPP}. 
In general, interactive path planning can be divided into two phases. 
In the first phase, a grid map is created and used to find multiple path candidates with a modified \ac{rrt} algorithm. 
This algorithm is able to find feasible paths in real-time and has been adapted to work without a preset target point. 
It favors smooth paths that do not cross lane markings.
The next step reduces the number of paths by applying a clustering algorithm such as k-means~\cite{Hosseini2014a} or DBSCAN~\cite{Schitz2021interactivePathPlanning}. 
The best path (e.g., in terms of smoothness) from each path cluster is chosen and suggested to the operator.
This gives the operator one path from each cluster to choose from.
In the second phase, the path chosen by the operator is further optimized to avoid obstacles and improve smoothness while maintaining driveability. 
Optimization can be conducted using the potential field method \cite{Hosseini2014a} or CHOMP~\cite{Schitz2021interactivePathPlanning}. 
Finally, the optimized path is then automatically tracked by the \ac{av}.

The level of abstraction reduction of this teleoperation concept comes at the cost of requiring a functional set of \ac{ad} modules such as perception, trajectory planning and following, which, on the other hand, effectively reduces the operator's workload, thus providing the main benefit of this concept.

\subsection{Perception Modification}
\label{sec:percMod}

\begin{figure}[!h]
	\centering
	\includegraphics[width=80mm]{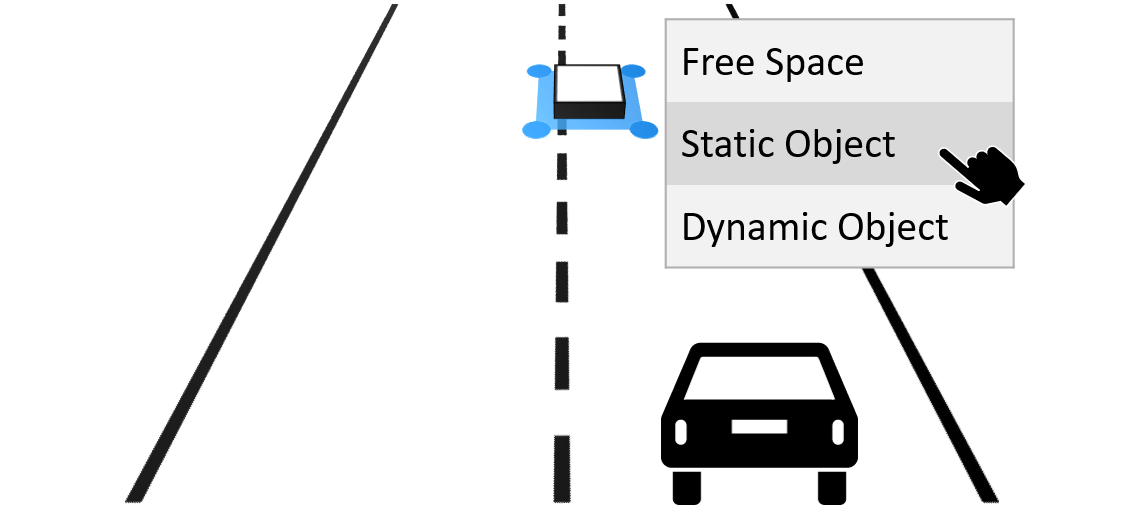}
	\caption[PercMod]{Perception modification concept. A human operator interacts with the perception module of the \ac{ad} function to resolve indeterminate object detections.
	}
	\label{fig:PercMod}
\end{figure}

With the objective of supporting the perception module of the \ac{ad} pipeline, Feiler and Diermeyer~\cite{Feiler2021percmod} proposed a teleoperation concept for perception modification. 
Conceptually visualized in Fig.~\ref{fig:PercMod}, this approach enables the resolving of situations in which either a false positive or an indeterminate and neglectable object detection hinders the \ac{av} operation. 
In this case, perception data, i.e., an object list and a grid map, are transmitted to a human operator and visualized in a 3D environment. 
With a video overlay, the human operator assesses the situation and draws a polygon around the  object with mouse clicks and keyboard inputs. 
Once approved, the information is sent to the vehicle and the \ac{ad} function classifies that area either as a free space or an obstacle. 
If the approval frees the desired path, the \ac{av} can continue its operation.

As described, this approach does not replace but enables interaction with the perception module of the \ac{ad} pipeline. 
It resolves perception-related fail cases of the \ac{ad} function and can be seen as a complement to the aforementioned concepts of shared control, waypoint guidance and interactive path planning, which all assume use of a perception module. 
Since the operator is not part of any planning or control task, it is assumed that the overall workload is lower that that of other teleoperation concepts.

%% file: Tables/overview_table_v2.tex
\begin{table}[!htpb]
	\caption{Overview of Teleoperation Concepts covered by the Survey}
	\label{tab:contrConcepts}
	\centering
	\begin{tblr}{
			stretch=1.75,
			colspec={Q[c,0.5cm]|Q[c,1.8cm]|Q[c,1.9cm]|Q[c,1.5cm]},
			rowspec={Q[m]Q[m]Q[m]Q[m]Q[m]Q[m]Q[m]Q[t]},
			vline{5,5} = {1-7}{1pt},
			vline{1,1} = {2-7}{1pt},
			vline{2,2} = {1-7}{1pt}}
		\SetHline[1]{2-4}{1pt}
		& \textbf{Concept} & \textbf{Publications} & \textbf{Input Devices} \\\hline[1pt]
		\SetCell[r=3,c=1]{c}{\rotatebox[origin=c]{90}{\textsc{Remote Driving}}} & Direct Control & \cite{Gnatzig2013ToDSystemDesign, Hofbauer2020telecarla, Schimpe21oss4tod, Jatzkowski2021, 5GCroCoD2p2, Appelqvist2007, Bensoussan1997, Bodell2016, Ross2007, Shen2016, Wu} & SW\&P \\ \hline
		& Shared Control & \cite{Schimpe2020steerwithme, Anderson2013, Qiao2021,  Schitz2021acc, Saparia21ass4tod, Storms2017sharedControl4oa} & SW\&P \\ \hline
		& Trajectory Guidance & \cite{Jatzkowski2021, Gnatzig2012trajBasedSharedAutonomy, Hoffmann2022a, Kay1995a} & SW\&P, M\&K \\ \hline[1pt]
		\SetCell[r=3,c=1]{c}{\rotatebox[origin=c]{90}{\textsc{Remote Assistance}}} & Waypoint Guidance & \cite{Schitz2021, 5GCroCoD2p2, Bjornberg2020} & M\&K \\ \hline
		& Interactive Path Planning & \cite{Hosseini2014a, Schitz2021interactivePathPlanning} & M\&K \\ \hline
		& Perception Modification   & \cite{Feiler2021percmod} & M\&K \\ \hline[1pt]
		\SetCell[r=1,c=4]{l}  SW\&P - Steering Wheel \& Pedals, \, M\&K - Mouse and Keyboard
	\end{tblr}
	\vspace*{-6mm}
\end{table}

%% file: Chapters/Patents.tex
\section{\uppercase{Industry Activities Relating to Vehicle Teleoperation}}
\label{sec:patents}
In addition to academic research, companies are also active in the area of \ac{av} teleoperation. 
Fig.~\ref{fig:patentChart} shows the number of patents published globally over the last 21 years. 
The data are based on results for the search term `Vehicle Teleoperation', obtained through the Google Patents search engine.
This terminology proved to filter out patents that are not directly related to automotive technology. 
The search was limited to include the filings in English for the time period from the year 2000 to 2021.
The exponential growth in the number of published patents is apparent.

\input{Figures/patentsFigure}
For a long time, remote control of \acp{av} was reserved for specific operational scenarios, usually in locations that would be dangerous and hazardous for manual operation by humans. 
However, finally it became evident that no matter how good full autonomy is, self-driving vehicles will always require some kind of remote monitoring, assistance, or driving to overcome challenging traffic scenarios. In consequence, teleoperation is being accepted as a viable fallback solution. 
This has not only motivated existing companies to increase their research and development activities in this field, but also motivated new startups to emerge thereby helping industry to scale up. 
In this context, Zoox became one of the most prominent companies at the forefront of research with numerous patents, e.g.,~\cite{Gogna2021a} and~\cite{Lockwood2021} as well as public demonstrations. 
A good example can be seen in the video~\cite{Zoox2020TeleGuidance} in which Zoox demonstrates its vision of waypoint guidance and perception modification on the streets of San Francisco. 
In contrast, Ottopia showcased both direct control and interactive path planning concepts as part of their Tel Aviv demonstration~\cite{OttopiaTechnologies2019} and joined forces with Motional to integrate teleoperation support into \acp{av}~\cite{Motional2022}. 
Volkswagen published a number of teleoperation-related patents, e.g.,~\cite{Rech2018} and has only recently announced teleoperation field-test activities with the German startup Fernride~\cite{fernride}. 
Furthermore, DriveU and EasyMile are collaborating to bring teleoperation to different \ac{av} types. 
Also worthy of mention are a number of other companies such as Argo AI, Aurora, Baidu, BMW, Robert Bosch, Cruise, Einride, Mercedes-Benz, Nissan, Phantom Auto, Renault, Vay, and others who are making significant advances in the field of \ac{av} teleoperation technology.

%% file: Figures/patentsFigure.tex
\newlength{\figurewidth}
\setlength{\figurewidth}{0.8\linewidth} 
\pgfplotsset{compat=newest}

\input{Figures/patent_variables}

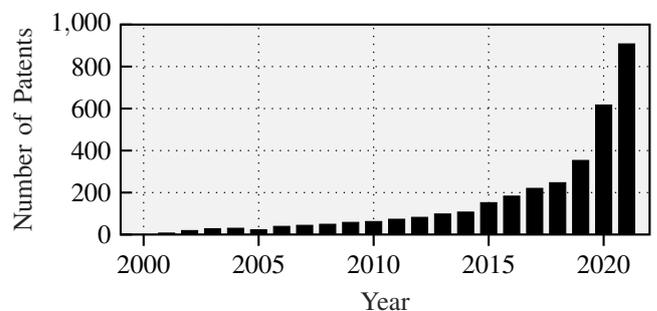
\begin{figure}[b]
	\centering
	\input{Figures/patentsChart}
	\caption{Published patents on "Vehicle Teleoperation" between years \patentsfirstYear\ and \patentslastYear\protect\footnotemark.}
	\label{fig:patentChart}
\end{figure}
\footnotetext{Data accessed on \patentsdate\ at \patentsurl\ }

%% file: Figures/patent_variables.tex
\newcommand\patentsdate{\text{08/05/22}} 
\newcommand\patentsfirstYear{\text{2000}} 
\newcommand\patentslastYear{\text{2021}} 
\newcommand\patentsurl{\url{https://patents.google.com/?q=\%28vehicle+teleoperation\%29\&before=priority:20211231\&after=priority:20000101\&language=ENGLISH}}

%% file: Figures/patentsChart.tex
%
%
\begin{tikzpicture}
	
\pgfplotsset{
	grid style = {
		dash pattern = on 0.05mm off 1mm,
		line cap = round,
		black,
		line width = 0.5pt
	}
}

\begin{axis}[%
width=1.0\figurewidth,
height=0.4\figurewidth,
at={(0\figurewidth,0\figurewidth)},
scale only axis,
bar shift auto,
xmin=1999,
xmax=2022,
xlabel style={font=\color{white!15!black}},
xlabel={Year},
ymin=0,
ymax=1000,
grid=both,
ylabel style={font=\color{white!15!black}},
ylabel={Number of Patents},
xtick align=outside,
xtick style={thick, black},
xtick pos=bottom,
ytick style={thick, black},
ytick pos=left,
axis background/.style={fill=white!95!black},
xticklabel style={/pgf/number format/1000 sep=}
]
\addplot[ybar, bar width=0.7, fill=black, area legend] table[row sep=crcr] {%
2001	7\\
2002	19\\
2003	28\\
2004	30\\
2005	23\\
2006	39\\
2007	44\\
2008	49\\
2009	58\\
2010	62\\
2011	73\\
2012	82\\
2013	98\\
2014	107\\
2015	152\\
2016	183\\
2017	220\\
2018	247\\
2019	353\\
2020	617\\
2021	908\\
};
\addplot[forget plot, color=white!15!black] table[row sep=crcr] {%
1999	0\\
2022	0\\
};
\end{axis}
\draw[black, thick] (0,0) rectangle (1.0\figurewidth,0.4\figurewidth);
\end{tikzpicture}%

%% file: Chapters/Conclusion.tex
\section{\uppercase{Conclusion}}
\label{sec:conclusion}
This paper has presented a systematic review of the literature on different teleoperation concepts and given an insight into their adoption in the \ac{av} industry. 
The concepts were grouped and discussed in terms of their applicability as fallback options with respect to the \ac{ad} pipeline.
In summary, it can be concluded that there is a wide variety of teleoperation concepts that can successfully handle different fail cases or \ac{odd} limitations within the \ac{ad} pipeline.
However, not all of them have the same significance, and, given their different functionalities, benefits and disadvantages, they will coexist in the future and continue to offer robust fallback solutions to \acp{av} with new concepts and features yet to come.

%
%

%% file: Chapters/Acknowledgements.tex
\section*{\uppercase{Acknowledgements}}
Domagoj Majstorovi\'c, Simon Hoffmann, Florian Pfab, Andreas Schimpe and Maria-Magdalena Wolf, as the first authors, collectively carried out the literature survey presented in this paper. 
Frank Diermeyer made essential contributions to the conception of the research projects on road vehicle teleoperation and revised the paper critically for important intellectual content. 
He gave final approval for the version to be published and agrees to all aspects of the work. 
As a guarantor, he accepts responsibility for the overall integrity of the paper.
The research was partially funded by the European Union~(EU) under RIA grant No.~825050, the Federal Ministry of Education and Research of Germany~(BMBF) within the project UNICARagil~(FKZ~16EMO0288), the Central Innovation Program~(ZIM) under grant No.~ZF4648101MS8, the project Wies'n Shuttle~(FKZ~03ZU1105AA) in the MCube cluster, and through basic research funds from the Institute for Automotive Technology.

%% file: root.bbl
\begin{thebibliography}{10}
\providecommand{\url}[1]{#1}
\csname url@rmstyle\endcsname
\providecommand{\newblock}{\relax}
\providecommand{\bibinfo}[2]{#2}
\providecommand\BIBentrySTDinterwordspacing{\spaceskip=0pt\relax}
\providecommand\BIBentryALTinterwordstretchfactor{4}
\providecommand\BIBentryALTinterwordspacing{\spaceskip=\fontdimen2\font plus
\BIBentryALTinterwordstretchfactor\fontdimen3\font minus
  \fontdimen4\font\relax}
\providecommand\BIBforeignlanguage[2]{{%
\expandafter\ifx\csname l@#1\endcsname\relax
\typeout{** WARNING: IEEEtran.bst: No hyphenation pattern has been}%
\typeout{** loaded for the language `#1'. Using the pattern for}%
\typeout{** the default language instead.}%
\else
\language=\csname l@#1\endcsname
\fi
#2}}

\bibitem{Guo2019hierarchicalCoopCtrl}
\BIBentryALTinterwordspacing
C.~Guo, C.~Sentouh, J.-B. Hau{\'{e}}, and J.-C. Popieul, ``{Driver–vehicle
  cooperation: a hierarchical cooperative control architecture for automated
  driving systems},'' \emph{Cognition, Technology {\&} Work}, vol.~21, no.~4,
  pp. 657--670, 2019.
\BIBentrySTDinterwordspacing

\bibitem{Walch2021}
\BIBentryALTinterwordspacing
M.~Walch, ``{Driver-Vehicle Interaction in Automated Driving : Overcoming
  System Boundaries via Driver Involvement},'' Ph.D. dissertation, Ulm
  University, 2021. [Online]. Available:
  \url{https://oparu.uni-ulm.de/xmlui/handle/123456789/40509}
\BIBentrySTDinterwordspacing

\bibitem{Bogdoll2021}
\BIBentryALTinterwordspacing
D.~Bogdoll, S.~Orf, L.~T{\"{o}}ttel, and J.~M. Z{\"{o}}llner, ``{Taxonomy and
  Survey on Remote Human Input Systems for Driving Automation Systems},''
  2022, pp. 94--108. [Online]. Available:
  \url{https://link.springer.com/10.1007/978-3-030-98015-3{\_}6}
\BIBentrySTDinterwordspacing

\bibitem{SAE2021}
\BIBentryALTinterwordspacing
{SAE J3016}, ``{J3016 - Taxonomy and Definitions for Terms Related to Driving
  Automation Systems for On-Road Motor Vehicles},'' Tech. Rep., 2021. [Online].
  Available: \url{https://www.sae.org/standards/content/j3016{\_}202104}
\BIBentrySTDinterwordspacing

\bibitem{CentreforConnected&AutomatedVehicles2022}
{Centre for Connected {\&} Automated Vehicles}, ``{Connected and automated
  vehicles - Vocabulary},'' BSI Flex 1890 v4.0:2022-03, 2022, Technical Report.

\bibitem{CentreforConnectedandAutonomousVehicles2020}
{Centre for Connected and Autonomous Vehicles}, ``{Guidelines for developing
  and assessing control systems for automated vehicles},'' BSI PAS 1880:2,
  2020, Technical Report.

\bibitem{Bensoussan1997}
\BIBentryALTinterwordspacing
S.~Bensoussan and M.~Parent, ``{Computer-aided teleoperation of an urban
  vehicle},'' in \emph{1997 8th International Conference on Advanced Robotics.
  Proceedings. ICAR'97}.\hskip 1em plus 0.5em minus 0.4em\relax IEEE, 1997, pp.
  787--792.
\BIBentrySTDinterwordspacing

\bibitem{Appelqvist2007}
\BIBentryALTinterwordspacing
P.~Appelqvist, J.~Knuuttila, and J.~Ahtiainen, ``{Development of an Unmanned
  Ground Vehicle for task-oriented operation - considerations on teleoperation
  and delay},'' in \emph{2007 IEEE/ASME international conference on advanced
  intelligent mechatronics}, 2007.
\BIBentrySTDinterwordspacing

\bibitem{Gnatzig2013ToDSystemDesign}
\BIBentryALTinterwordspacing
S.~Gnatzig, F.~Chucholowski, T.~Tang, and M.~Lienkamp, ``{A System Design for
  Teleoperated Road Vehicles},'' in \emph{Proceedings of the 10th International
  Conference on Informatics in Control, Automation and Robotics}.\hskip 1em
  plus 0.5em minus 0.4em\relax SciTePress - Science and and Technology
  Publications, 2013, pp. 231--238.
\BIBentrySTDinterwordspacing

\bibitem{Shen2016}
\BIBentryALTinterwordspacing
X.~Shen, Z.~J. Chong, S.~Pendleton, G.~M. {James Fu}, B.~Qin, E.~Frazzoli, and
  M.~H. Ang, ``{Teleoperation of On-Road Vehicles via Immersive Telepresence
  Using Off-the-shelf Components},'' in \emph{Advances in Intelligent Systems
  and Computing}, 2016, pp. 1419--1433. [Online]. Available:
  \url{http://link.springer.com/10.1007/978-3-319-08338-4{\_}102}
\BIBentrySTDinterwordspacing

\bibitem{Hofbauer2020telecarla}
\BIBentryALTinterwordspacing
M.~Hofbauer, C.~B. Kuhn, G.~Petrovic, and E.~Steinbach, ``{TELECARLA: An Open
  Source Extension of the CARLA Simulator for Teleoperated Driving Research
  Using Off-the-Shelf Components},'' in \emph{2020 IEEE Intelligent Vehicles
  Symposium (IV)}.\hskip 1em plus 0.5em minus 0.4em\relax IEEE,  2020, pp.
  335--340.
\BIBentrySTDinterwordspacing

\bibitem{Schimpe21oss4tod}
\BIBentryALTinterwordspacing
A.~Schimpe, J.~Feiler, S.~Hoffmann, D.~Majstorovic, and F.~Diermeyer, ``{Open
  Source Software for Teleoperated Driving},'' in \emph{2022 International
  Conference on Connected Vehicle and Expo (ICCVE)}.\hskip 1em plus 0.5em minus
  0.4em\relax IEEE,  2022, pp. 1--6.
\BIBentrySTDinterwordspacing

\bibitem{Jatzkowski2021}
I.~Jatzkowski, T.~Stolte, R.~Graubohm, and P.~M. Maurer, ``{Integration of a
  Vehicle Operating Mode Management into UNICAR agil's Automotive
  Service-oriented Software Architecture},'' \emph{30th Aachen Colloquium
  Sustainable Mobility 2021}, pp. 595--614, 2021.

\bibitem{5GCroCoD2p2}
5GCroCo, ``{Deliverable D2.2: Test Case Definition and Test Site Description
  for Second Round Tests and Trials},'' Fifth Generation Cross-Border Control,
  Tech. Rep. 825050, 2021.

\bibitem{Ross2007}
\BIBentryALTinterwordspacing
B.~Ross, J.~Bares, D.~Stager, L.~Jackel, and M.~Perschbacher, ``{An Advanced
  Teleoperation Testbed},'' in \emph{Field and Service Robotics}.\hskip 1em
  plus 0.5em minus 0.4em\relax Berlin, Heidelberg: Springer Berlin Heidelberg,
  2008, vol.~42, pp. 297--304. [Online]. Available:
  \url{http://link.springer.com/10.1007/978-3-540-75404-6{\_}28}
\BIBentrySTDinterwordspacing

\bibitem{Bodell2016}
O.~Bodell and E.~Gulliksson, ``{Teleoperation of Autonomous Vehicle With 360°
  Camera Feedback},'' Master's thesis in Systems, Control and Mechatronics,
  2016.

\bibitem{Wu}
Y.-C. Wu, ``{The Implementation of Remote Driving for Autonomous Vehicle over
  the LTE Wireless Network},'' pp. 3--5.

\bibitem{Anderson2013}
\BIBentryALTinterwordspacing
S.~J. Anderson, S.~B. Karumanchi, K.~Iagnemma, and J.~M. Walker, ``{The
  intelligent copilot: A constraint-based approach to shared-adaptive control
  of ground vehicles},'' \emph{IEEE Intelligent Transportation Systems
  Magazine}, vol.~5, no.~2, pp. 45--54, 2013.
\BIBentrySTDinterwordspacing

\bibitem{Schimpe2020steerwithme}
\BIBentryALTinterwordspacing
A.~Schimpe and F.~Diermeyer, ``{Steer with Me: A Predictive, Potential
  Field-Based Control Approach for Semi-Autonomous, Teleoperated Road
  Vehicles},'' in \emph{2020 IEEE 23rd International Conference on Intelligent
  Transportation Systems (ITSC)}.\hskip 1em plus 0.5em minus 0.4em\relax IEEE,
  2020, pp. 1--6.
\BIBentrySTDinterwordspacing

\bibitem{Qiao2021}
\BIBentryALTinterwordspacing
B.~Qiao, H.~Li, and X.~Wu, ``{Intelligent-Assist Algorithm for Remote
  Shared-Control Driving Based on Game Theory},'' \emph{Journal of Shanghai
  Jiaotong University (Science)}, vol.~26, no.~5, pp. 615--625, 2021.
\BIBentrySTDinterwordspacing

\bibitem{Schitz2021acc}
\BIBentryALTinterwordspacing
D.~Schitz, G.~Graf, D.~Rieth, and H.~Aschemann, ``{Model-Predictive Cruise
  Control for Direct Teleoperated Driving Tasks},'' in \emph{2021 European
  Control Conference, ECC 2021}.\hskip 1em plus 0.5em minus 0.4em\relax IEEE,
  2021, pp. 1808--1813.
\BIBentrySTDinterwordspacing

\bibitem{Storms2017sharedControl4oa}
\BIBentryALTinterwordspacing
J.~Storms, K.~Chen, and D.~Tilbury, ``{A shared control method for obstacle
  avoidance with mobile robots and its interaction with communication delay},''
  \emph{The International Journal of Robotics Research}, vol.~36, no. 5-7, pp.
  820--839, 2017.
\BIBentrySTDinterwordspacing

\bibitem{Saparia21ass4tod}
\BIBentryALTinterwordspacing
S.~Saparia, A.~Schimpe, and L.~Ferranti, ``{Active Safety System for
  Semi-Autonomous Teleoperated Vehicles},'' in \emph{2021 IEEE Intelligent
  Vehicles Symposium Workshops (IV Workshops)}.\hskip 1em plus 0.5em minus
  0.4em\relax IEEE,  2021, pp. 141--147.
\BIBentrySTDinterwordspacing

\bibitem{Gnatzig2012trajBasedSharedAutonomy}
\BIBentryALTinterwordspacing
S.~Gnatzig, F.~Schuller, and M.~Lienkamp, ``{Human-machine interaction as key
  technology for driverless driving - A trajectory-based shared autonomy
  control approach},'' in \emph{2012 IEEE RO-MAN: The 21st IEEE International
  Symposium on Robot and Human Interactive Communication}.\hskip 1em plus 0.5em
  minus 0.4em\relax IEEE,  2012, pp. 913--918.
\BIBentrySTDinterwordspacing

\bibitem{Hoffmann2022a}
\BIBentryALTinterwordspacing
S.~Hoffmann, D.~Majstorovic, and F.~Diermeyer, ``{Safe Corridor: A
  Trajectory-Based Safety Concept for Teleoperated Road Vehicles},'' in
  \emph{2022 International Conference on Connected Vehicle and Expo
  (ICCVE)}.\hskip 1em plus 0.5em minus 0.4em\relax IEEE,  2022, pp. 1--6.
\BIBentrySTDinterwordspacing

\bibitem{Kay1995a}
\BIBentryALTinterwordspacing
J.~S. Kay and C.~E. Thorpe, ``{Operator Interface Design Issues in a
  Low-Bandwidth and High-Latency Vehicle Teleoperation System},'' in \emph{SAE
  Technical Papers},  1995.
\BIBentrySTDinterwordspacing

\bibitem{Bjornberg2020}
A.~Bj{\"{o}}rnberg, ``{Shared Control for Vehicle Teleoperation with a Virtual
  Environment Interface},'' Master's thesis in Systems, Control and Robotics,
  2020.

\bibitem{Schitz2021}
\BIBentryALTinterwordspacing
D.~Schitz, G.~Graf, D.~Rieth, and H.~Aschemann, ``{Interactive Corridor-Based
  Path Planning for Teleoperated Driving},'' in \emph{2021 7th International
  Conference on Mechatronics and Robotics Engineering (ICMRE)}.\hskip 1em plus
  0.5em minus 0.4em\relax IEEE,  2021, pp. 174--179.
\BIBentrySTDinterwordspacing

\bibitem{Hosseini2014a}
\BIBentryALTinterwordspacing
A.~Hosseini, T.~Wiedemann, and M.~Lienkamp, ``{Interactive path planning for
  teleoperated road vehicles in urban environments},'' in \emph{17th
  International IEEE Conference on Intelligent Transportation Systems
  (ITSC)}.\hskip 1em plus 0.5em minus 0.4em\relax IEEE,  2014, pp. 400--405.
\BIBentrySTDinterwordspacing

\bibitem{Schitz2021interactivePathPlanning}
\BIBentryALTinterwordspacing
D.~Schitz, S.~Bao, D.~Rieth, and H.~Aschemann, ``{Shared Autonomy for
  Teleoperated Driving: A Real-Time Interactive Path Planning Approach},'' in
  \emph{2021 IEEE International Conference on Robotics and Automation (ICRA)},
  vol. 2021-May.\hskip 1em plus 0.5em minus 0.4em\relax IEEE,  2021, pp.
  999--1004.
\BIBentrySTDinterwordspacing

\bibitem{Feiler2021percmod}
\BIBentryALTinterwordspacing
J.~Feiler and F.~Diermeyer, ``{The Perception Modification Concept to Free the
  Path of An Automated Vehicle Remotely},'' in \emph{Proceedings of the 7th
  International Conference on Vehicle Technology and Intelligent Transport
  Systems}.\hskip 1em plus 0.5em minus 0.4em\relax SCITEPRESS - Science and
  Technology Publications, 2021, pp. 405--412.
\BIBentrySTDinterwordspacing

\bibitem{Georg2020}
\BIBentryALTinterwordspacing
J.-M. Georg, J.~Feiler, S.~Hoffmann, and F.~Diermeyer, ``{Sensor and Actuator
  Latency during Teleoperation of Automated Vehicles},'' in \emph{2020 IEEE
  Intelligent Vehicles Symposium (IV)}.\hskip 1em plus 0.5em minus 0.4em\relax
  IEEE,  2020, pp. 760--766.
\BIBentrySTDinterwordspacing

\bibitem{Mutzenich2021}
\BIBentryALTinterwordspacing
C.~Mutzenich, S.~Durant, S.~Helman, and P.~Dalton, ``{Updating our
  understanding of situation awareness in relation to remote operators of
  autonomous vehicles},'' \emph{Cognitive Research: Principles and
  Implications}, vol.~6, no.~1, p.~9, 2021.
\BIBentrySTDinterwordspacing

\bibitem{Hoffmann2021safetyAssessment}
\BIBentryALTinterwordspacing
S.~Hoffmann and F.~Diermeyer, ``{Systems-theoretic Safety Assessment of
  Teleoperated Road Vehicles},'' in \emph{Proceedings of the 7th International
  Conference on Vehicle Technology and Intelligent Transport Systems}.\hskip
  1em plus 0.5em minus 0.4em\relax SCITEPRESS - Science and Technology
  Publications, 2021, pp. 446--456.
\BIBentrySTDinterwordspacing

\bibitem{Chucholowski2013a}
\BIBentryALTinterwordspacing
F.~E. Chucholowski, ``{Evaluation of Display Methods for Teleoperation of Road
  Vehicles},'' \emph{Journal of Unmanned System Technology}, vol.~3, no.~3, pp.
  80--85, 2016.
\BIBentrySTDinterwordspacing

\bibitem{Tang2014c}
\BIBentryALTinterwordspacing
T.~Tang, P.~Vetter, S.~Finkl, K.~Figel, and M.~Lienkamp, ``{Teleoperated Road
  Vehicles – The "Free Corridor" as a Safety Strategy Approach},''
  \emph{Applied Mechanics and Materials}, vol. 490-491, pp. 1399--1409, 2014.
\BIBentrySTDinterwordspacing

\bibitem{Graf2020a}
G.~Graf, Y.~Abdelrahman, H.~Xu, Y.~Abdrabou, D.~Schitz, H.~Hu{\ss}mann, and
  F.~Alt, ``{The Predictive Corridor : A Virtual Augmented Driving Assistance
  System for Teleoperated Autonomous Vehicles},'' in \emph{International
  Conference on Artificial Reality and Telexistence and Eurographics Symposium
  on Virtual Environments (ICAT-EGVE)}, 2020.

\bibitem{Georg2019a}
\BIBentryALTinterwordspacing
J.-M. Georg and F.~Diermeyer, ``{An Adaptable and Immersive Real Time Interface
  for Resolving System Limitations of Automated Vehicles with Teleoperation},''
  in \emph{2019 IEEE International Conference on Systems, Man and Cybernetics
  (SMC)}, vol. 2019-Octob.\hskip 1em plus 0.5em minus 0.4em\relax IEEE,  2019,
  pp. 2659--2664.
\BIBentrySTDinterwordspacing

\bibitem{Georgb}
\BIBentryALTinterwordspacing
J.-M. Georg, E.~Putz, and F.~Diermeyer, ``{Longtime Effects of Videoquality,
  Videocanvases and Displays on Situation Awareness during Teleoperation of
  Automated Vehicles},'' in \emph{2020 IEEE International Conference on
  Systems, Man, and Cybernetics (SMC)}, vol. 2020-Octob.\hskip 1em plus 0.5em
  minus 0.4em\relax IEEE,  2020, pp. 248--255.
\BIBentrySTDinterwordspacing

\bibitem{Georg2018}
\BIBentryALTinterwordspacing
J.-M. Georg, J.~Feiler, F.~Diermeyer, and M.~Lienkamp, ``{Teleoperated Driving,
  a Key Technology for Automated Driving? Comparison of Actual Test Drives with
  a Head Mounted Display and Conventional Monitors},'' in \emph{2018 21st
  International Conference on Intelligent Transportation Systems (ITSC)}.\hskip
  1em plus 0.5em minus 0.4em\relax IEEE,  2018, pp. 3403--3408.
\BIBentrySTDinterwordspacing

\bibitem{Bout2017}
\BIBentryALTinterwordspacing
M.~Bout, A.~P. Brenden, M.~Klingeg{\aa}rd, A.~Habibovic, and M.-P.
  B{\"{o}}ckle, ``{A Head-Mounted Display to Support Teleoperations of Shared
  Automated Vehicles},'' in \emph{Proceedings of the 9th International
  Conference on Automotive User Interfaces and Interactive Vehicular
  Applications Adjunct}.\hskip 1em plus 0.5em minus 0.4em\relax New York, NY,
  USA: ACM,  2017, pp. 62--66.
\BIBentrySTDinterwordspacing

\bibitem{Hosseini2016e}
\BIBentryALTinterwordspacing
A.~Hosseini and M.~Lienkamp, ``{Predictive safety based on track-before-detect
  for teleoperated driving through communication time delay},'' in \emph{2016
  IEEE Intelligent Vehicles Symposium (IV)}.\hskip 1em plus 0.5em minus
  0.4em\relax IEEE,  2016, pp. 165--172.
\BIBentrySTDinterwordspacing

\bibitem{Hosseini2016a}
\BIBentryALTinterwordspacing
A.~Hosseini, F.~Richthammer, and M.~Lienkamp, ``{Predictive Haptic Feedback for
  Safe Lateral Control of Teleoperated Road Vehicles in Urban Areas},'' in
  \emph{2016 IEEE 83rd Vehicular Technology Conference (VTC Spring)}.\hskip 1em
  plus 0.5em minus 0.4em\relax IEEE,  2016, pp. 1--7.
\BIBentrySTDinterwordspacing

\bibitem{Gogna2021a}
R.~Gogna, ``{Teleoperations for Collaborative Vehicle Guidance},'' Patent,
  2021, WO2021211322A1.

\bibitem{Lockwood2021}
A.~L.~K. Lockwood, R.~Gogna, G.~Linscott, P.~Orecchio, D.~Xie, A.~G. Rege, and
  J.~S. Levinson, ``{Predictive teleoperator situational awareness},'' Patent,
  2019, US10976732B2.

\bibitem{Zoox2020TeleGuidance}
\BIBentryALTinterwordspacing
Zoox, ``{How Zoox Uses TeleGuidance to Provide Remote Assistance to its
  Autonomous Vehicles},'' 2022. [Online]. Available:
  \url{https://youtu.be/NKQHuutVx78}
\BIBentrySTDinterwordspacing

\bibitem{OttopiaTechnologies2019}
\BIBentryALTinterwordspacing
{Ottopia Technologies}, ``{How Ottopia uses differente Remote Assistance
  Techniques to operate Autonomous Vehicles},'' 2019. [Online]. Available:
  \url{https://youtu.be/IIkLSwaTJg8}
\BIBentrySTDinterwordspacing

\bibitem{Motional2022}
\BIBentryALTinterwordspacing
Motional, ``{Motional's Remote Vehicle Assistance (RVA)},'' 2022. [Online].
  Available: \url{https://youtu.be/pyoHeEcgHFA}
\BIBentrySTDinterwordspacing

\bibitem{Rech2018}
B.~Rech, S.~Gl{\"{a}}ser, M.~Engel, H.-J. G{\"{u}}nther, T.~Buburuzan,
  S.~Kleinau, B.~Lehmann, and J.~Hartog, ``{Method for remotely controlling a
  plurality of driverless self-driving systems, control station for remotely
  controlling the self-driving systems, and system},'' Patent, 2018,
  WO2018171991A1.

\bibitem{fernride}
\BIBentryALTinterwordspacing
Handelsblatt, ``{VW und Fernride testen in Wolfsburg ferngesteuerte Lkw},''
  2022. [Online]. Available:
  \url{https://www.handelsblatt.com/technik/it-internet/automatisiertes-fahren-vw-und-fernride-testen-in-wolfsburg-ferngesteuerte-lkw/28278722.html}
\BIBentrySTDinterwordspacing

\end{thebibliography}
